\documentclass{article}
\usepackage{spconf,amsmath,graphicx, url}
\usepackage[colorlinks,linkcolor=red]{hyperref}

\title{Cross-Modal Supervised Learning For Better Acoustic Representations}
\name{Shaoyong Jia, Xin Shu, Yang Yang, Dawei Liang, Qiyue Liu, Junhui Liu}
\address{iQiYi, Inc. Beijing, China \\ \small{jiashaoyong@qiyi.com}}
\begin{document}
%
\maketitle
\begin{abstract}
Obtaining large-scale human-labeled datasets to train acoustic representation models is a very challenging task. On the contrary, we can easily collect data with machine-generated labels. In this work, we propose to exploit machine-generated labels to learn better acoustic representations, based on the synchronization between vision and audio. Firstly, we collect a large-scale video dataset with 15 million samples, which totally last 16,320 hours. Each video is 3 to 5 seconds in length and annotated automatically by publicly available visual and audio classification models. Secondly, we train various classical convolutional neural networks (CNNs) including VGGish\cite{hershey2017cnn}, ResNet\_50\cite{he2016identity} and Mobilenet\_v2\cite{sandler2018mobilenetv2}. We also make several improvements to VGGish and achieve better results. Finally, we transfer our models on three external standard benchmarks for audio classification task, and achieve significant performance boost over the state-of-the-art results. Models and codes are available at: \url{https://github.com/Deeperjia/vgg-like-audio-models}.
\end{abstract}
\section{Introduction}
\label{sec:intro}

As a popular research area in audio community, audio classification is widely applied in wildlife monitoring\cite{briggs2012acoustic}, context recognition\cite{eronen2006audio}, etc. Traditionally, low-level audio feature descriptors such as filter banks\cite{beltran2015scalable}, MFCC\cite{phan2015representing} and time-frequency descriptors\cite{chu2009environmental} are utilized followed with Gaussian Mixture Models, Bag of Words\cite{lu2015sparse} and Support Vector Machines\cite{huang2013blind}. Recent works, such as VGGish\cite{hershey2017cnn} and SoundNet\cite{aytar2016soundnet}, are now developing acoustic representations models with CNN, because of its end-to-end training ability and great success in the vision community.

Most approaches to audio classification depend on high-quality datasets for supervised learning. Previous datasets, such as LITIS\cite{rakotomamonjy2015histogram}, ESC \cite{stowell2015detection}, DCASE\cite{piczak2015esc}, TUT\cite{mesaros2018multi}, are limited in size, which makes it hard to train models with large capacity. Recently, Audio Set\cite{gemmeke2017audio} with  2.1 million 10-second clips from YouTube and 527 sample-level audio event labels, is released. However, some labels are weak since they sometimes occur in a time window of 1-3 seconds. Youtube-100M\cite{hershey2017cnn} is another large-scale but private internal dataset of 100 million samples with 30,871 labels, which are collected in the wild. However, it is weakly annotated because labels are machine generated and assigned to the entrie video with an average duration of 4.6 minutes, which leads to confusion in frame-level supervised learning. Moreover, it costs large amounts of computing resources and much time for training. 

\begin{figure}[htb]
	\includegraphics[width=\linewidth]{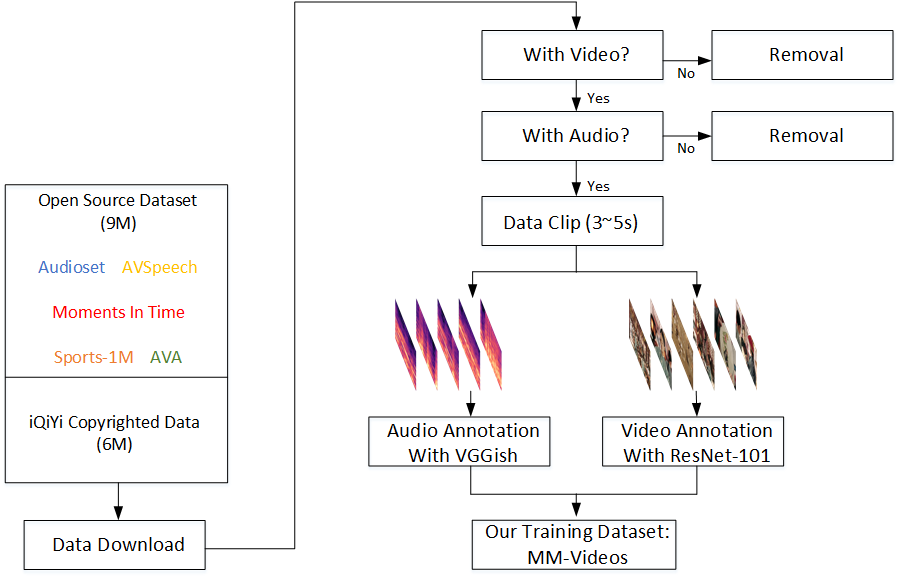}
	\caption{The main process of collecting videos.}\label{Fig.1}
\end{figure}

In this paper, we firstly propose an efficient method to build a video dataset only for training acoustic models with collecting data mainly from open-source datasets, as illustrated in Fig.\ref{Fig.1}. We collect 15 million videos (16,320 hours), each tagged automatically from a set of 10,998 multimodal labels coming from ML-Images\cite{wu2019tencent} and Audio Set\cite{gemmeke2017audio}. We call it \emph{MM-Videos}. The average duration of videos is 4 seconds, which is fine-grained and corresponds to the average duration of human working memory\cite{barrouillet2004time}. Then we train various CNN models, including VGGish, ResNet\_50 and Mobilenet\_v2. We establish an improved architecture based on VGGish. To our knowledge, we are the first to publish competitive performance of Mobilenet\_v2\cite{sandler2018mobilenetv2} applied to audio.

The similar method of transferring vision to audio is SoundNet\cite{aytar2016soundnet}, which utilizes the synchronous nature of vision and audio. There are three significant differences. Firstly, all the labels come from visual in SoundNet while we also tag each video with audio labels. SoundNet\cite{aytar2016soundnet} uses 1,401 visual categories coming from Imagenet\cite{russakovsky2015imagenet} and Palces2\cite{zhou2015places2} while we use 10,988 multi-modal labels. It is demonstrated in \cite{hershey2017cnn} that pre-trained models on the dataset with a larger number of various categories usually shows better performance. Secondly, SoundNet learns the distribution of probabilities while our audio models directly learn from the labels annotated by visual, which decrease the storage space and simplify the calculation during training. Thirdly, audios are sent into one-dimensional CNN directly in SoundNet while we extracts two-dimensional log-Mel spectrogram through short time Fourier transformation (STFT) for two-dimensional CNN models.

\section{Dataset}
\label{sec:dataset}

{\bf Data Collection}: We collect videos from various famous open datasets, including Moments in Time\cite{monfort2019moments}, Audio Set\cite{gemmeke2017audio}, AV-Speech\cite{ephrat2018looking}, AVA\cite{murray2012ava}, Sports-1M\cite{karpathy2014large}. We ignore the labels in the datasets. As a supplement, iQiYi copyrighted videos covering all channel types are randomly merged into \emph{MM-Videos}. We reserve the videos, which have both vision and audio components. Note that we only use the unbalanced and balanced training subsets in Audio Set\cite{gemmeke2017audio}.  We cut each video into 3 to 5 second-long clips. 

{\bf Data Annotation}: The multi-modal category vocabulary consisting of 11,166 vision labels from ML-Images\cite{wu2019tencent} and 527 audio labels from Audio Set\cite{gemmeke2017audio}. We annotate each video with visual and audio CNN models by machine, respectively. We sample 3 frames per second and it results in 9 to 15 frames per video. Each frame is sent to ResNet\_101\cite{wu2019tencent}, which is trained on ML-Images, to infer visual labels. As for audio, log-Mel spectrograms are extracted through STFT and sent to VGGish, to infer labels for each 1-second non-overlapping audio segment. VGGish is a VGG-like audio classification model that is pre-trained on YouTube-100M\cite{hershey2017cnn} and fine-tuned on Audio Set\cite{gemmeke2017audio}. Video-level labels are obtained as the average of frame-level predictions. After averaging, top-10 predictions are selected as the visual labels. We filter the average predictions value with top-5 and a fixed threshold to generate 3 auido labels on average. In summary,
10,988 categories appear during annotation among the total 11,693
labels and the utilization rate of vocabulary is 93.97\%. The rest 695
labels are abandoned since they do not appear in our dataset.

\begin{figure}[htb]

\begin{minipage}[b]{1.0\linewidth}
  \centering
  \centerline{\includegraphics[width=8.5cm]{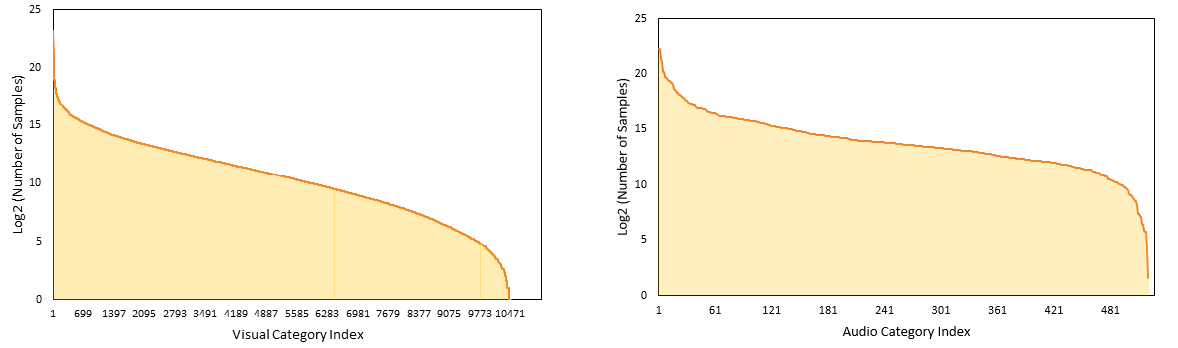}}
\end{minipage}
\begin{minipage}[b]{1.0\linewidth}
  \centering
  \centerline{\includegraphics[width=8.5cm]{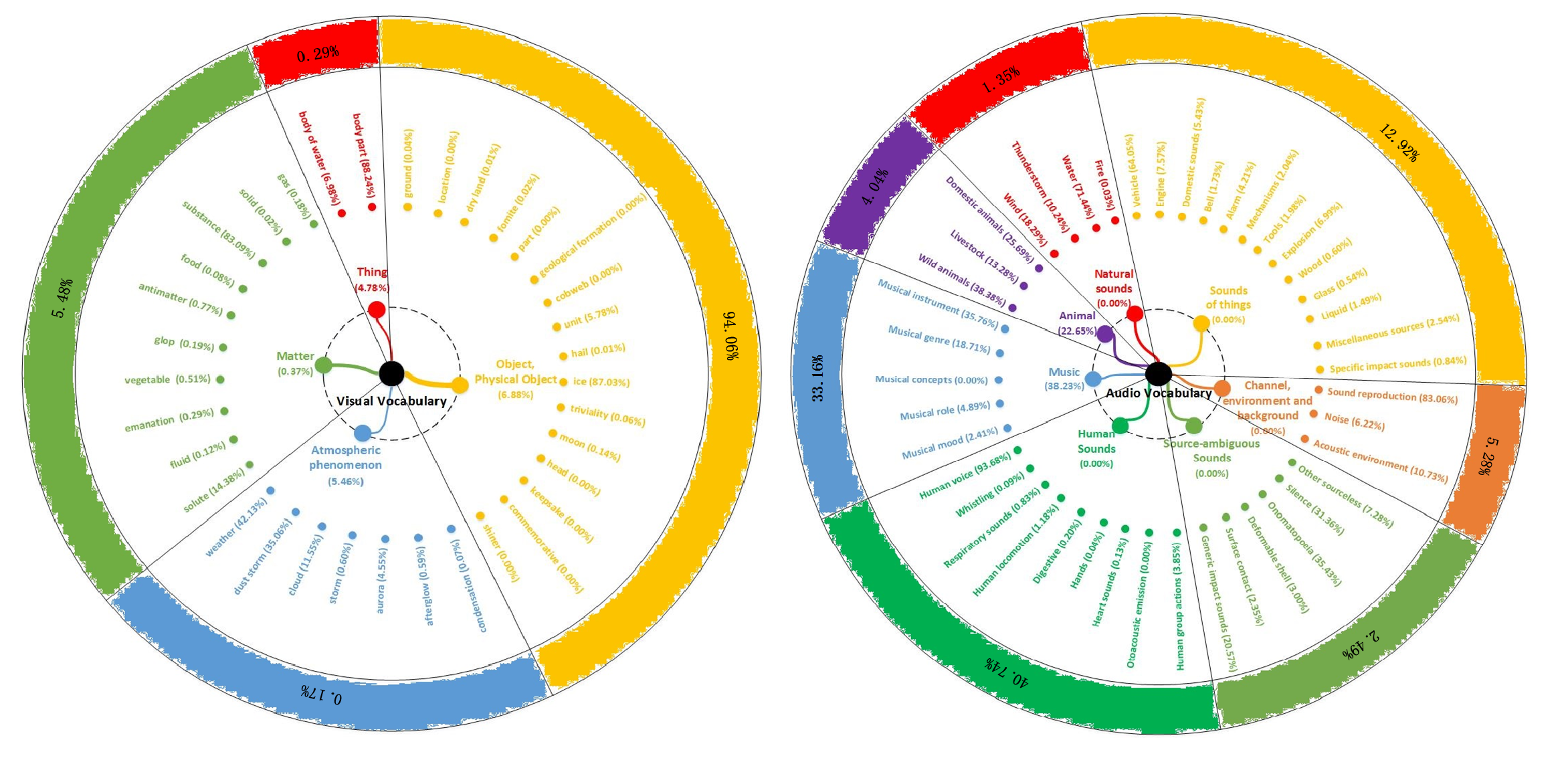}}
\end{minipage}
\caption{It is better to view by zooming in three times at least. (top) Number of samples per category in visual vocabulary and audio vocabulary in \emph{MM-Videos}. The number of samples for each category is calculated by log2 and they all fall in the range of 0 to 25. (bottom) Distribution of categories in visual and audio vocabulary. Since the vocabulary satisfies the tree structure, proportions in the black color stand for how much the sum of subtrees of this primary category (including itself, as a father node) occupying in the whole amount of samples. While proportions in other colors expound how much the sum of subtrees of this secondary category (including itself, as a father node) or the primary category (as an independent node) occupying in the sum of subtrees of the corresponding primary category.}
\label{Fig.2}
\end{figure}

{\bf Dataset Statistics}: \emph{MM-Videos} has 15 million videos and 10,998 multi-modal labels with a duration of 3 to 5 seconds per video. It is split into 14 million training subset and 1 million validation subset. Each video is annotated with about 13 labels, 10 from visual vocabulary and approximately 3 from the audio vocabulary. It is obvious that imbalance among different kinds of categories appears in both vision and audio, which can be seen in Fig.\ref{Fig.2}. We divide them into 4,096 different folds by Universally Unique Identifier (UUID) Hash. Compared with Youtube-100M, \emph{MM-Videos} is only 0.3\% of Youtube-100M for training models and shows better acoustic representations performance, demonstrated in Section 4.

\section{CNN Design and Architecture}
\label{sec:exper_frame}

{\bf Input Features}: The audio is first converted to single-channel at a 16kHz sampling rate and is divided into non-overlapping 1000ms frames. Each frame inherits all the labels of its parent video. Spectrograms are then generated in a sliding window fashion using a hamming window of width 25ms and step 10ms. Each spectrogram is integrated into 64 mel-spaced frequency bins, and the magnitude of each bin is log-transformed after adding a small offset to avoid numerical issues. This gives log-Mel spectrogram of 100*64 size as input to the CNN models.

{\bf Architecture}: Several CNN models are trained for acoustic representations learning. The VGGish model is trained with different kinds of configurations. We also train deeper CNN models, ResNet\_50\cite{he2016identity} with bottleneck for a stronger ability to feature expression and MobileNet\_v2\cite{sandler2018mobilenetv2} with competitive performance for mobile and embedded applications. And some changes are applied to these models to adapt to train on \emph{MM-Videos}. We use a bottleneck layer of 128 units placed before the final output layer to accelerate training. If necessary, the stride of the convolutional layer is decreased by half to guarantee an output feature map of 6*4. As \emph{MM-Videos} is a multi-label training dataset, we use a sigmoid layer of 10,988 units as a classifier. To our knowledge, we are the first to publish results of Mobilenet\_v2\cite{sandler2018mobilenetv2} applied to audio. 

VGGish\cite{hershey2017cnn} is a classical VGG-like model widely used in audio classification. We repalce the fully-connected layers, including the one resized to 128 embedding, by convolutional layers. Besides, batch normalization is added after each activation layer in convolutional layers to accelerate the convergence of training and improve the acoustic representations performance. This designed architecture of our improved VGGish is shown in Table~\ref{tab:vggish}, and it has approximately 360.72 million FLOPs and 73.54 million parameters.

{\bf Loss Function}: In multi-label classification, standard cross-entropy function after sigmoid is used as the loss function. The cross-entropy function on the sample \begin{math}x_{i}\end{math} and the ground truth label \begin{math}y_{i}\end{math} is calculated as follows:

\begin{equation}
  Loss(x_{i},y_{i})=\frac{1}{m}\sum_{j=0}^m [-y_{i,j}log(p_{i,j})-(1-y_{i,j})log(1-p_{i,j})]
\end{equation}
where \begin{math}p_{i,j}\end{math} donates the posterior probability with respect to the category \begin{math}j\end{math} and \begin{math}m\end{math} donates the size of category vocabulary.

\begin{table}
	\centering
	\caption{The network details of our improved VGGish model.}
	\label{tab:vggish}
	\begin{tabular}{cc} 
		\hline
		\multicolumn{2}{c}{The Architecture of Our VGGish Model}  \\ 
		\hline
		Layer               & Kernel Size                         \\ 
		\hline
		Convolution with BN & 3*3*64                              \\ 
		\hline
		Max Pooling         & 2*2                                 \\ 
		\hline
		Convolution with BN & 3*3*128                             \\ 
		\hline
		Max Pooling         & 2*2                                 \\ 
		\hline
		Convolution with BN & {[}3*3, 256]*2                       \\ 
		\hline
		Max Pooling         & 2*2                                 \\ 
		\hline
		Convolution with BN & {[}3*3, 512]*2                       \\ 
		\hline
		Max Pooling         & 2*2                                 \\ 
		\hline
		Convolution with BN & 6*4*4096                            \\ 
		\hline
		Convolution with BN & 1*1*4096                            \\ 
		\hline
		Convolution with BN & 1*1*128                             \\ 
		\hline
		\multicolumn{2}{c}{Fully-connected-10998}                 \\ 
		\hline
		\multicolumn{2}{c}{Sigmoid}                               \\
		\hline
	\end{tabular}
\end{table}

\begin{table}[htb]
	\centering
	\caption{Performances of different pre-trained models on Audio Set\cite{gemmeke2017audio}. For our pre-trained "VGGish" lines, the changes are cumulative and each subsequent line includes the new change on the previous ones. Except for ResNet\_50\cite{hershey2017cnn}, all the models are finetuned  only on balanced training subset. }
	\label{tab:eval_audioset}
	\begin{tabular}{cccccc} 
		\hline
		\multicolumn{5}{c}{Performances of Different models on Audio Set}                                                    \\ 
		\hline
		Models                                                                  & Top-1  & Top-5   & mAP   & AUC    \\ 
		\hline
		ResNet\_50\cite{hershey2017cnn}                                         & -      & -      & 31.4\% & 95.9\%  \\ 
		\hline
		\begin{tabular}[c]{@{}c@{}}VGGish\\Youtube-100M\cite{hershey2017cnn} \end{tabular}           & 55.3\% & 74.9\% & 23.1\% & 94.3\%  \\ 
		\hline\hline
		\begin{tabular}[c]{@{}c@{}}VGGish\\\emph{MM-Videos} \end{tabular}          & 58.3\% & 78.1\% & 27.0\% & 95.0\%  \\ 
		\hline
		\begin{tabular}[c]{@{}c@{}}VGGish\\Batch Normalization \end{tabular}    & 59.1\% & 78.5\% & 28.6\% & 95.2\%  \\ 
		\hline
		\begin{tabular}[c]{@{}c@{}}VGGish\\Fully Convolution \end{tabular}      & 59.9\% & 79.4\% & 30.6\% & 95.5\%  \\ 
		\hline
		MobileNet\_v2                                                           & 55.1\% & 74.3\% & 21.8\% & 93.7\%  \\
		\hline
		ResNet\_50                                                         & \textbf{60.9\%} & \textbf{81.1\%} & \textbf{32.3\%} & \textbf{96.0\%}  \\ 
		\hline
	\end{tabular}
\end{table}

{\bf Training}: To reduce overfitting, we augment the data by taking random 1-second non-overlapping segment in the time domain for each video. As for labels, every video in the \emph{MM-Videos} has 10 visual labels and 1 to 5 audio labels. Half of the visual labels are selected randomly to increase randomness and all the audio labels are directly applied as the training labels. Thus, a set of 6 to 10 multi-modal labels corresponds to each 1-second video segment. 

All CNN models are trained asynchronously on 4 GPUs of \emph{Nvidia K40} within 14 days for about 30 epochs using Adam\cite{kingma2014adam} optimizer. The batch-size is 1,024 for VGGish-like and Resnet\_50 models and 2,048 for Mobilenet\_v2. Learning rate is initialized to 0.0001 and it decays with a factor of 0.9 after every epoch. Besides, \emph{L2} regularization is added to the loss function to avoid overfitting and the \emph{L2} regularization penalty is set to 1.0. During training, we saw no evidence of overfitting by monitoring progress via 1-best accuracy and mean Average Precision (mAP) over a validation subset.

\section{Experiments}
\label{sec:eval_exper}

{\bf Evaluation Benchmarks}: Classifier models using embeddings from our acoustic representations are applied on standard benchmarks ESC-50\cite{stowell2015detection}, TUT Acoustic Scene 2018\cite{mesaros2018multi} and Audio Set\cite{gemmeke2017audio}, to evaluate the performance.

\begin{table}[htb]
	\centering
	\caption{Performances of different pre-trained models on ESC-50\cite{stowell2015detection}. For our pre-trained "VGGish" lines, the changes are cumulative and each subsequent line includes the new change on the previous ones.}
	\label{tab:eval_esc}
	\begin{tabular}{ccc} 
		\hline
		\multicolumn{3}{c}{Performances of Different models on ESC-50}                            \\ 
		\hline
		Models                                                                 & Top-1  & mAP    \\ 
		\hline
		Human Performance\cite{stowell2015detection}                           & 81.3\% & -      \\ 
		\hline
		FBEs+ConvRBM-BANK\cite{sailor2017unsupervised}                         & 86.5\% & -      \\ 
		\hline
		\begin{tabular}[c]{@{}c@{}}VGGish\\Youtube-100M\cite{hershey2017cnn}\end{tabular}           & 81.1\% & 86.1\%  \\ 
		\hline\hline
		\begin{tabular}[c]{@{}c@{}}VGGish\\\emph{MM-Videos}\end{tabular}          & 87.3\% & 91.7\%  \\ 
		\hline
		\begin{tabular}[c]{@{}c@{}}VGGish\\Batch Normalization\end{tabular}    & 90.3\% & 95.2\%  \\ 
		\hline
		\begin{tabular}[c]{@{}c@{}}VGGish\\Fully Convolution\end{tabular}      & \textbf{91.4\%} & \textbf{95.9\%} \\  
		\hline
		MobileNet\_v2                                                          & 83.8\% & 89.4\%  \\
		\hline
		ResNet\_50                                                         & 91.2\% & 95.5\%  \\ 
		\hline
	\end{tabular}
\end{table}

\begin{table}
	\centering
	\caption{Performances of different pre-trained models on TUT Acoustic Scene 2018\cite{mesaros2018multi}. For our pre-trained "VGGish" lines, the changes are cumulative and each subsequent line includes the new change on the previous ones.}
	\label{tab:eval_tut}
	\begin{tabular}{ccc} 
		\hline
		\multicolumn{3}{c}{Performance of Different models on TUT Acoustic Scene 2018}               \\ 
		\hline
		Models                                                                  & Top-1   & mAP      \\ 
		\hline
		baseline\cite{mesaros2018multi}                                         & 59.7\% & -        \\ 
		\hline
		\begin{tabular}[c]{@{}c@{}}VGGish\\Youtube-100M\cite{hershey2017cnn} \end{tabular}           & 62.6\% & 67.0\%  \\ 
		\hline\hline
		\begin{tabular}[c]{@{}c@{}}VGGish\\\emph{MM-Videos} \end{tabular}          & 68.2\% & 71.6\%  \\ 
		\hline
		\begin{tabular}[c]{@{}c@{}}VGGish\\Batch Normalization \end{tabular}    & 70.1\% & 75.2\%  \\ 
		\hline
		\begin{tabular}[c]{@{}c@{}}VGGish\\Fully Convolution \end{tabular}      & \textbf{71.4\%} & \textbf{76.8\%}  \\ 
		\hline
		MobileNet\_v2                                                           & 66.2\% & 69.5\%  \\ 
		\hline
		ResNet\_50                                                          & 70.7\% & 75.3\%  \\
		\hline
	\end{tabular}
\end{table}

ESC-50 is a balanced dataset containing 2,000 5-second audio samples with 50 labels. The data is rearranged into 5 folds and the accuracy results are reported as the mean of five leave-one-fold-out evaluations. The development dataset of TUT Acoustic Scene 2018\footnote{http://dcase.community/challenge2018/task-acoustic-scene-classification} contains 10 acoustic scenes. There are 8,640 10-second audio segments, 6,122 segments in training set and 2,518 segments in the evaluation set. They both are small-scale, single-label and manually-annotated balanced datasets. Audio Set is a much larger and multi-label imbalanced dataset of 2.1 million 10-second samples with 527 audio event classes. Audio Set is divided in three disjoint subsets: a balanced evaluation subset, a balanced training subset, and an unbalanced training subset. The classifier model is trained only on the balanced subset.

{\bf Transfer Learning}: To train classfier models using embeddings from our acoustic representations, the last layer of 10,998 units is alternated with the adapted class vocabulary size of the benchmark. We concat the 128-dimensional audio embedding features as the inputs for classifier models.

The learning rate is 0.0002, batch size is set to 128 and L2 regularization weight decay is set to $10^{-6}$. Besides, dropout varies from 0.3 for ESC-50\cite{stowell2015detection} to 0.5 for Audio Set\cite{gemmeke2017audio} and TUT Acoustic Scene 2018\cite{mesaros2018multi} to reduce overfitting.

{\bf Evaluation Metrics}: We use four metrics (Top-1, Top-5, mAP, AUC) to evaluate the performance of different acoustic representations models. Top-n accuracy is the fraction of test samples that contain at least one of the ground truth labels in the top-n predictions. mAP is the mean across categories of the Average Precision. AUC is the area under the Receiver Operating Characteristic curve.

{\bf Results}: We evaluate seven groups of models, including five groups of VGGish\cite{hershey2017cnn}, one group of ResNet\_50\cite{he2016identity} and MobileNet\_v2\cite{sandler2018mobilenetv2} with the width multiplier of 1.0. On all benchmarks, the VGGish pre-trained on \emph{MM-Videos} outperforms the same model pre-trained on Youtube-100M\cite{hershey2017cnn} by at least 3\% of both TOP-1 and mAP, which proves our effective way to collect data for cross-modal deep training for acoustic models with the size only around 0.3\% of Youtube-100M.

More analysis on Audio Set, batch normalization and using convolutional layer instead of the fully-connected layer also lead to an increase of 0.8\% in Top-1 and at least 1.6\% in mAP, respectively. Importantly, we finetune our ResNet\_50 model only on balanced training subset and get best Top-1 of 60.9\% and mAP of 32.3\% , which exceeds the Google's baseline by 0.9\% in mAP of ResNet\_50, which is finetuned on unbalanced and balanced training subsets. Almost the same conclusions can be obtained on Top-5, and AUC, which are used for large-scale and multi-label dataset of Audio Set.

The comparison experiments of different models on ESC-50 and TUT Acoustic Scene 2018 reflect similar conclusions as those on Audio Set, by getting more obvious better performance. But the performance of VGGish with fully convolution get best performance, which mainly caused by the small-scale characteristic. The homepage of ESC-50\footnote{https://github.com/karoldvl/ESC-50} enumerates large amounts of solutions. Specifically, the human can recognize with Top-1 of 81.3\% and the unsupervised filterbank learning using convolutional restricted Boltzmann machine\cite{sailor2017unsupervised} achieves the best result of 86.5\%. The fully convolution VGGish model reaches 91.4\% Top-1 accuracy which has an increase of almost 5\%. There are some methods which maybe achieve better performance on TUT Acoustic Scene 2018 since they utilize multi model fusion algorithm or multi complex features, such as the temporal information, binaural audio inputs or larger log-Mel spectrogram size.

\section{Conclusions}
We collect 15 million videos, which last 16,320 hours in total, with fine-grained automatic annotation multi-modal sample-level labels mainly from open datasets. Besides, an improved architecture on the classical VGGish model is proposed and the importance of each part of the changes are illustrated through ablation analysis. We train this improved VGGish, classical VGGish, MobileNet\_v2, and ResNet\_50 to learn the acoustic representations. We have demonstrated the value of collecting data for cross-modal weakly supervised learning for acoustic representations, by evaluation on the Audio Set, ESC-50, and TUT Acoustic Scene 2018 standard downstream benchmarks for audio classification task with better performance. The success of our method is a significant step towards acoustic
representation learning since it shows robustness on different benchmarks with obviously reducing the dataset size and time complexity of model training.

\vfill\pagebreak

\bibliographystyle{IEEEbib}
\bibliography{refs}

\end{document}